\documentclass[11pt]{article}

\usepackage[utf8]{inputenc}
\usepackage[margin=1in]{geometry}
\usepackage{amsmath,amssymb}
\usepackage{graphicx}
\usepackage{booktabs}
\usepackage{multirow}
\usepackage{array}
\usepackage{hyperref}
\usepackage{authblk}
\usepackage[numbers]{natbib}  

\newcommand{\revise}[1]{#1}   

\title{A Statistical Approach for Modeling Irregular Multivariate Time Series with Missing Observations}

\author[1]{Dingyi Nie}
\author[1]{Yixing Wu\thanks{Corresponding author}}
\author[1]{C.-C. Jay Kuo}
\affil[1]{University of Southern California}

\date{}

\date{}  

\begin{document}

\maketitle

\begin{abstract}
Irregular multivariate time series with missing values present significant challenges for predictive modeling in domains such as healthcare. While deep learning approaches often focus on temporal interpolation or complex architectures to handle irregularities, we propose a simpler yet effective alternative: extracting time-agnostic summary statistics to eliminate the temporal axis. Our method computes four key features per variable-mean and standard deviation of observed values, as well as the mean and variability of changes between consecutive observations to create a fixed-dimensional representation. These features are then utilized with standard classifiers, such as logistic regression and XGBoost. Evaluated on four biomedical datasets (PhysioNet Challenge 2012, 2019, PAMAP2, and MIMIC-III), our approach achieves state-of-the-art performance, surpassing recent transformer and graph-based models by 0.5-1.7\% in AUROC/AUPRC and 1.1-1.7\% in accuracy/F1-score, while reducing computational complexity. Ablation studies demonstrate that feature extraction-not classifier choice-drives performance gains, and our summary statistics outperform raw/imputed input in most benchmarks. In particular, we identify scenarios where missing patterns themselves encode predictive signals, as in sepsis prediction (PhysioNet, 2019), where missing indicators alone can achieve 94.2\% AUROC with XGBoost, only 1.6\% lower than using original raw data as input. Our results challenge the necessity of complex temporal modeling when task objectives permit time-agnostic representations, providing an efficient and interpretable solution for irregular time series classification.
\end{abstract}

\section{Introduction}


Multivariate time series represent a critical data modality in numerous scientific domains and are fundamental for modeling real-world problems. The past decade has witnessed the emergence of various deep neural network architectures capable of processing multi-dimensional sequential data with variable lengths. These include temporal convolutional neural networks (TCN)\cite{lea2017temporal}, various variants of recurrent neural networks (RNN) \cite{chung2014empirical, koutnik2014clockwork, gers2000learning}, and Transformer-based models\cite{vaswani2017attention, wen2022transformers}. Although these approaches excel with well-defined and fully observed sequential data, real-world time series often present significant challenges. First, sampling frequently occurs at irregular intervals, where identical observed values with different time gaps between consecutive points can carry substantially different meanings. Second, missing values are common in time series data, resulting from either the inherent nature of the studied phenomenon or uncontrollable factors in data collection procedures. A good example of such challenges is clinical data. Vital signals are typically captured by sensors that are vulnerable to power outages or technical failures, creating gaps in continuous monitoring. Meanwhile, clinicians order lab tests at their discretion, based on their clinical assessment of the patient's condition and medical necessity, resulting in inherently irregular sampling patterns.

Although existing methods can be adapted for irregular multivariate time series with missing observations—by discarding missing values, applying kernel- or interpolation-based techniques\cite{schafer2002missing, lu2008reproducing}, or implementing temporal warping and resampling\cite{desautels2016prediction, moor2019early}—these approaches remain suboptimal, as they overlook potentially informative missing patterns\cite{che2018recurrent, agniel2018biases, little2019statistical}. To address these challenges, machine learning researchers have made attempts in recent years to develop specialized neural network architectures that not only accommodate missing values and irregular sampling but also leverage these apparent deficiencies as valuable information sources.

To name a few:
\begin{itemize}
    \item \textbf{GRU-D}\cite{che2018recurrent}: a modified version of the gated recurrent unit (GRU)\cite{cho2014learning, chung2014empirical} that involves two learnable decay terms in the GRU's input and hidden state paths. It also takes as input the concatenation of three components: multivariate time series variables, observed versus missing masking labels, and time intervals since the last observation, at each recurrent step.
    \item \textbf{SeFT}\cite{horn2020set}: end-to-end learning based on attention\cite{vaswani2017attention} of set functions that map multivariate time series encoded as sets to classification outcomes.
    \item \textbf{mTAND}\cite{shukla2021multi}: takes irregularly sampled time points and variable values as attention keys and values to produce a fixed-dimensional representation at the time points of the query.
    \item \textbf{Raindrop}\cite{zhang2021graph}: a representation based on a graph neural network (GNN)\cite{wu2020comprehensive} of irregular multivariate time series with missing values, estimating the latent sensor graph structure and utilizing nearby observations to predict misaligned readouts.
    \item \textbf{ViTST}\cite{li2024time}: visualizes time series using integrated line graphs and treats time series classification as image classification using vision transformers\cite{dosovitskiy2020image, liu2021swin}.
\end{itemize}

Although these deep learning studies continue to improve benchmark accuracies in biomedical and clinical time series classification tasks, a survey of critical care research reproducibility demonstrates that gradient boosting trees often achieve results comparable to deep learning methods in popular public critical care datasets \cite{johnson2017reproducibility}. Another study shows that tree-based models remain state-of-the-art in medium-sized tabular data compared to deep learning models \cite{grinsztajn2022tree, wu2023discriminant}. Advanced gradient boosting methods such as XGBoost \cite{chen2016xgboost} efficiently capture key statistical relationships between data and target variables, delivering strong performance in many biomedical modeling tasks \cite{wu2025prediction, wang2025green} while inherently handling missing values. Considering the complex architectures, substantial computational requirements, and challenging training and hyperparameter optimization processes of deep learning methods, we are motivated to develop a more straightforward yet practical approach for modeling irregular multivariate time series with missing observations.

In this study, we propose a two-step pipeline that extracts key statistics from irregular multivariate time series with missing observations, creating statistical features as representations, followed by classification using logistic regression or gradient boosting. Our feature extraction approach eliminates the temporal axis of the time series, making it versatile and adaptable to the \revise{datasets with diverse characteristics (e.g., varying numbers of variables, sequence lengths, and sample sizes)} and missing data rates. We evaluate our method by comparing its classification performance with state-of-the-art deep learning baselines across four popular biomedical and clinical time series classification benchmark datasets. The results demonstrate that our approach achieves leading accuracies while offering significantly reduced computational complexity and streamlined implementation for both training and inference. We also conducted extensive experiments by comparing our statistical features with those of the raw input and imputed time series using different imputation strategies, followed by the same classifier. We also conducted extensive experiments, comparing our statistical features against both raw input and time series data processed with various imputation strategies, all of which were evaluated using identical classifiers. The results demonstrate that our method's superior performance stems not solely from the classifier (XGBoost) but significantly from our feature extraction process, which substantially enhances classification effectiveness. In general, our study introduces a practical approach to modeling irregular multivariate time series with missing observations, challenging the notion that complex architectures are necessary for this task.

\section{Methods}
Our proposed methods involve two stages: feature extraction and classification.

\subsection{Feature Extraction}
Let a multivariate time series segment be represented as $\boldsymbol{X} = \{ \boldsymbol{x}_{i} \}_{i=1}^L \in \mathbb{R}^{L \times D}$, where $L$ denotes the length of the segment (number of time steps) and $D$ is the number of variables. The corresponding timestamps $\boldsymbol{t} = \{ t_i \}_{i=1}^L$ are in a consistent unit (e.g., hours or minutes). At each time step $i$, multiple observations of different variables that occur simultaneously are consolidated into the same vector $\boldsymbol{x}_i = [x_{i,1},\dots,x_{i,D}] \in \mathbb{R}^D$, which may contain the initial missing values. To explicitly tell the real observations and missing values apart, we introduce a masking array $\boldsymbol{M} = \{\boldsymbol{m}_i\} \in \{0, 1\}^{L \times D}$ of the same shape as $\boldsymbol{X}$, where
\begin{equation}
{m}_{i,d} = \begin{cases} 
      1, & \text{if } {x}_{i,d} \text{ is a real observation,} \\
      0, & \text{otherwise.} 
   \end{cases}
\end{equation}

To characterize each variable independently of time, we compute the following statistical features.

\textbf{Mean of Observed Values}: For each variable \( d \), we calculate the mean of its observed values, denoted as:
\begin{equation}
\mu^{(0)}_d = \frac{1}{\sum_{i=1}^L m_{i,d}} \sum_{i=1}^L m_{i,d} \, x_{i,d}.
\end{equation}

\textbf{Standard Deviation of Observed Values}: To quantify the spread of observed values, we calculate the standard deviation for each variable:
\begin{equation}
\sigma^{(0)}_d = \sqrt{\frac{1}{\sum_{i=1}^L m_{i,d}} \sum_{i=1}^L m_{i,d} \, (x_{i,d} - \mu^{(0)}_d)^2}.
\end{equation}

To avoid division by zero, in case $\sum_{i=1}^L m_{i,d} = 0$ (i.e., no observed values for the variable $d$ in the segment), we replace $\mu^{(0)}_d$ with the \revise{global} average $\bar{\mu}^{(0)}_d$ \revise{(computed using all observed values of variable $d$ across the entire training set)}, and simply replace $\sigma^{(0)}_d$ with 0.

To further characterize the tendency of each variable to change over time, we compute two more statistical features that measure the changes in variable observations.

\textbf{Mean Change In Values}:  The mean change in values for variable $d$ is defined as
\begin{equation}
\mu^{(1)}_d = \frac{1}{\sum_{i=1}^L m_{i,d} - 1} \sum_{(i, j) \in \mathcal{P}_d} (x_{j,d} - x_{i,d}).
\end{equation}
where $\mathcal{P}_d$ denotes the set of all pairs $(i, j)$ such that $x_{i,d}$ and $x_{j,d}$ are \revise{temporally} consecutive observed values for variable $d$ \revise{(where $i < j$)}, and the number of such pairs $|\mathcal{P}_d| = \sum_{i=1}^L m_{i,d} - 1$.

\textbf{Standard Deviation of Change In Values}: To capture the variability in the rate of change, we estimate the standard deviation of the rates between consecutive observed values for each variable $d$:
\begin{equation}
\sigma^{(1)}_d = \sqrt{\frac{1}{\sum_{i=1}^L m_{i,d} - 1} \sum_{\{(i, j) \in \mathcal{P}_d\}} \left( (x_{j,d} - x_{i,d}) - \mu^{(1)}_d \right)^2}.
\end{equation}

If there are no consecutive observations for a variable (i.e., $\sum_{i=1}^L m_{i,d} \leq 1$), we replace $\mu^{(1)}_d$ and $\sigma^{(1)}_d$ with 0.

By representing each variable $d$ in one time series segment with these four statistical features, concatenated as a new feature vector
\begin{equation}
\boldsymbol{f}_d = \begin{bmatrix}
    \mu_d^{(0)} & \sigma_d^{(0)} & \mu_d^{(1)} & \sigma_d^{(1)}
\end{bmatrix}^\top \in \mathbb{R}^{4},
\end{equation}
we effectively transform $\mathbf{X} \in \mathbb{R}^{L \times D}$ into a fixed-size, time-independent representation $\boldsymbol{F} = \begin{bmatrix}
    \boldsymbol{f}_1 & \boldsymbol{f}_2 & \cdots & \boldsymbol{f}_D
\end{bmatrix} \in \mathbb{R}^{4 \times D}$, where temporal dependencies (time stamp array $\boldsymbol{t}$ and missing patterns in the missing array $\boldsymbol{M}$) are no longer required.

\subsection{Classification and Evaluation}
Once the feature array $\boldsymbol{F}$ is acquired, we build prediction models using gradient boosting (XGBoost) from \textit{xgboost (version 1.7.3, Python implementation)}, \revise{as well as logistic regression (LR), \revise{random forest (RF), and support vector machine (SVM)} from \textit{scikit-learn (version 1.4.2)} as classification task heads.} The feature array $\boldsymbol{F}$ is flattened and is used directly as input for these models. \revise{We used a consistent parameter setting for the XGBoost classifier across all experiments: 100 trees, a maximum depth of 5, and a learning rate of 0.1. We then employed 5-fold cross-validation to obtain robust performance metrics.} It is essential to note that for LR, the input feature array $\boldsymbol{F}$ is standardized before modeling, using scaling parameters calculated from the training fold during cross-validation.

In this study, we focus on classification tasks, where the prediction target can be either binary or multi-class, depending on the dataset being evaluated. For binary classification models, we use the area under the receiver operator characteristic curve (AUROC) as the evaluation metric. To account for label imbalance in the evaluation datasets, we also report the area under the precision-recall curve (AUPRC) as a complementary metric. For multi-class classification models, we use the accuracy, precision, recall, and F1 score, averaged across all classes, as the metric. We report the average metrics and their standard deviations across all folds to evaluate performance. For each dataset, we compare the performance of our models with that of a set of state-of-the-art deep learning and neural network-based methods by considering the metrics against those reported in the original study under similar experimental settings.

A detailed listing of datasets and the deep learning methods to be compared is presented in the results section.

\section{Results}


\subsection{Datasets}
Time series datasets, particularly those with irregular sampling and missing values, remain relatively underexplored in machine learning research, with only a few well-established benchmarks. Additionally, due to availability and privacy concerns, previous works often utilize different datasets and employ varying cohort selection criteria, making comparisons challenging. However, we have identified several studies from the deep learning community in recent years that exhibit similarities in dataset selection and task definition.

To evaluate our proposed method, we selected four representative real-world time series datasets, consistent with previous studies, for easier comparison. Among these datasets, three come with missing values (two of which are also irregularly sampled). At the same time, the fourth originally had no missing values but was later modified to include them manually. They are:

\paragraph*{PhysioNet Challenge 2019 (P19)}
The PhysioNet Challenge 2019 dataset focuses on the early prediction of sepsis using hourly-sampled clinical time series of varying lengths. It includes 34 dynamic variables, including vital signs such as \textit{HeartRate} and \textit{Temperature}, as well as laboratory test values such as \textit{pH} and \textit{Glucose}. We used data from both training sets A and B, totaling 40,336 time series segments. The original challenge task was to identify the exact time at which the patient enters the stage of sepsis. To align with existing studies, we assign binary labels to each segment as a whole, depending on whether sepsis has occurred or not, essentially transforming the task into a segment-level binary classification problem. 92.7\% of the segments are negative (non-sepsis). We do not use the six static demographic variables from the original dataset, as they remain constant over time. The average time series sequence length is 38, and the maximum sequence length is 336. P19 has a high level of missing data, with 26 out of 34 time series variables having a missing rate greater than 90\%.

\paragraph*{PhysioNet Challenge 2012 (P12)}
The PhysioNet Challenge 2012 dataset comprises irregularly sampled time series data collected during the first two days of a patient's stay in the intensive care unit (ICU), with multiple outcome descriptors. We utilize it for the binary classification task of predicting in-hospital mortality. After removing 12 outliers following SeFT\cite{horn2020set}, P12 has 11,988 time series instances in total, consisting of 37 variables (mainly laboratory test results) that change over time. We do not use the five static demographic variables associated with each patient in the dataset. 85.8\% of the samples are negative (survivors). The average time series sequence length is 75, with the maximum length being 215. Given its irregular sample rates, which vary across different variables, P12 also exhibits a high level of missing data. The \textit{HR} (heart rate) variable has the least missing rate of 23.4\%, while 22 of the 37 variables have a missing rate greater than 90\%.

\paragraph*{PAMAP2 Physical Activity Monitoring (PAM)}
The PAMAP2 dataset is used to recognize human activity and includes data collected from 9 subjects performing activities of daily living using three inertial measurement units (IMUs). Originally, PAMAP2 was sampled at a fixed rate, and all variables were fully observed. However, several recent studies from the deep learning community have adopted it for the task of modeling time series with missing values by manually dropping out partial observations. Here, we follow the preprocessing steps of Raindrop \ cite {zhang2021graph}, resulting in a set of 5,333 time series segments with eight physical activities as classification target labels. There are 17 time-dependent variables sampled at a fixed rate of 100 Hz. All segments have a sequence length of 600, and all variables are manually set to be 60\% missing.

\paragraph*{MIMIC-III}
MIMIC-III is an extensive biomedical database derived primarily from clinical records of ICU stays. In this study, we utilized the subset of in-hospital mortality prediction from the established public benchmark dataset \textit{MIMIC-III} \cite{Harutyunyan2019}. The dataset contains 17 irregularly sampled time-dependent variables extracted from the first 48 hours of a patient's ICU stay, and the target is to predict whether the patient will die in the hospital or not. This dataset comprises 21,139 time series instances in total, with 86.8\% of them being negative instances, i.e., in-hospital survivors. The average sequence length is 84, and the maximum length is 2,879. This dataset also exhibits high missing variable rates, with 10 of the 17 variables having a missing rate greater than 80\%, and 5 of them exceeding 90\%.


\subsection{Data Availability}
All four aforementioned datasets are available online. The two PhysioNet Challenge datasets can be found on the official PhysioNet website at \url{https://www.physionet.org/content/challenge-2019/} and \url{https://www.physionet.org/content/challenge-2012/}. PAMAP2 can be found publicly in the UC Irvine Machine Learning Repository at \url{https://archive.ics.uci.edu/dataset/231/pamap2+physical+activity+monitoring}. Access to the MIMIC-III Database can be requested on the PhysioNet website at \url{https://physionet.org/content/mimiciii/1.4/}, but prerequisite training must first be completed. The processing scripts that we used to create the in-hospital mortality benchmark subset can be found at \url{https://github.com/YerevaNN/mimic3-benchmarks}.

\subsection{Comparison to State-of-the-Art Deep Learning Methods}

We evaluate our summary feature-based method on all four biomedical time series datasets and compare its performance with some of the state-of-the-art deep learning methods for irregular time series modeling.

In P19, P12 and PAM, we compare our method with Transformer\cite{vaswani2017attention}, GRU-D\cite{che2018recurrent}, SeFT\cite{horn2020set}, mTAND\cite{shukla2021multi}, IP-Net\cite{shukla2019interpolation}, DGM$^2$-O\cite{wu2021dynamic}, MTGNN\cite{jin2022multivariate}, Raindrop\cite{zhang2021graph}, and ViTST\cite{li2024time}. The performance of these models was reported in ViTST, under settings similar to those used in our study. Table \ref{tab:p19-p12-pam_results} shows the evaluation results in $(mean\pm std)$ format. Please note that all the deep learning models we compare with are based on stochastic gradient descent; therefore, they are all nondeterministic algorithms, and the \textit{mean} and \textit{std} are calculated over multiple training and evaluation runs. In our case, we calculate the \textit{mean} and \textit{std} statistics from 5-fold cross-validation.

Our method, specifically with the XGBoost classifier, outperforms all existing approaches in three classification tasks: sepsis occurrence in the P19 dataset, mortality prediction in the P12 dataset, and human activity classification in the PAM dataset. Notably, our method demonstrates strong consistency, as reflected in the relatively low standard deviation of evaluation scores across different folds.
In the P19 and P12 datasets, our method achieves improvements of 0.5\% and 0.6\% in absolute AUROC, as well as 0.7\% and 1.2\% in absolute AUPRC, over the previous state-of-the-art deep learning model ViTST. For the PAM dataset, the performance gains are even more pronounced, with increases of 1.4\% in accuracy, 1.7\% in precision, 1.2\% in recall, and an additional 1.1\% in absolute F1 score. \revise{To validate further that the performance gains are primarily attributable to our extracted features rather than the choice of a specific classifier, we evaluated our approach using LR, RF, and SVM. As shown in Table~\ref{tab:p19-p12-pam_results} and Table~\ref{tab:mimic3_results}, all three classifiers achieved competitive performance across the datasets. Notably, Ours (RF) consistently matches or outperforms many deep learning baselines, such as achieving an AUROC of 88.2\% on P19 and an F1 score of 95.3\% on PAM. This consistency across diverse classifiers confirms that the discriminative power stems essentially from the proposed statistical features.}

\begin{table*}[ht]
\centering
{\fontsize{6}{9}\selectfont
\renewcommand{\arraystretch}{1}
\setlength{\tabcolsep}{1pt}
\begin{tabular}{l|cc|cc|cccc}
\hline
Methods & \multicolumn{2}{c|}{P19} & \multicolumn{2}{c|}{P12} & \multicolumn{4}{c}{PAM} \\ \cline{2-9} 
 & AUROC & AUPRC & AUROC & AUPRC & Accuracy & Precision & Recall & F1 score \\ \hline
Transformer & 80.7 $\pm$ 3.8 & 42.7 $\pm$ 7.7 & 83.3 $\pm$ 0.7 & 47.9 $\pm$ 3.6 & 83.5 $\pm$ 1.5 & 84.8 $\pm$ 1.5 & 86.0 $\pm$ 1.2 & 85.0 $\pm$ 1.3 \\
Trans-mean & 83.7 $\pm$ 1.8 & 45.8 $\pm$ 3.2 & 82.6 $\pm$ 2.0 & 46.3 $\pm$ 4.0 & 83.7 $\pm$ 2.3 & 84.9 $\pm$ 2.6 & 86.4 $\pm$ 2.1 & 85.4 $\pm$ 2.4 \\
GRU-D & 83.9 $\pm$ 1.7 & 46.9 $\pm$ 2.1 & 81.9 $\pm$ 2.1 & 46.1 $\pm$ 4.7 & 83.3 $\pm$ 1.6 & 84.6 $\pm$ 1.6 & 85.2 $\pm$ 1.3 & 84.8 $\pm$ 1.2 \\
SeFT & 81.2 $\pm$ 2.3 & 41.9 $\pm$ 3.1 & 73.9 $\pm$ 2.5 & 31.1 $\pm$ 1.4 & 67.1 $\pm$ 2.2 & 70.0 $\pm$ 2.4 & 68.2 $\pm$ 1.5 & 68.5 $\pm$ 1.8 \\
mTAND & 84.4 $\pm$ 1.3 & 50.6 $\pm$ 2.4 & 84.2 $\pm$ 0.8 & 48.2 $\pm$ 3.4 & 74.6 $\pm$ 4.3 & 74.3 $\pm$ 4.0 & 79.5 $\pm$ 2.8 & 76.8 $\pm$ 3.4 \\
IP-Net & 84.6 $\pm$ 1.3 & 38.1 $\pm$ 3.7 & 82.6 $\pm$ 1.4 & 47.6 $\pm$ 3.1 & 74.3 $\pm$ 3.8 & 75.6 $\pm$ 2.1 & 77.9 $\pm$ 2.2 & 76.6 $\pm$ 2.8 \\
DGM$^2$-O & 86.7 $\pm$ 3.4 & 44.7 $\pm$ 1.7 & 84.4 $\pm$ 1.6 & 47.3 $\pm$ 3.6 & 82.4 $\pm$ 2.3 & 85.2 $\pm$ 2.1 & 83.9 $\pm$ 2.3 & 84.3 $\pm$ 1.8 \\
MTGNN & 81.9 $\pm$ 6.2 & 39.9 $\pm$ 8.9 & 74.4 $\pm$ 6.7 & 35.5 $\pm$ 6.6 & 80.3 $\pm$ 4.1 & 85.2 $\pm$ 1.7 & 86.1 $\pm$ 1.9 & 85.9 $\pm$ 2.4 \\
Raindrop & 87.0 $\pm$ 2.3 & 51.8 $\pm$ 5.5 & 82.8 $\pm$ 1.7 & 44.0 $\pm$ 3.0 & 88.5 $\pm$ 1.5 & 89.9 $\pm$ 1.5 & 89.9 $\pm$ 0.6 & 89.8 $\pm$ 1.0 \\
ViTST & 89.2 $\pm$ 2.0 & 53.1 $\pm$ 3.4 & 85.1 $\pm$ 0.8 & 51.1 $\pm$ 4.1 & 95.8 $\pm$ 1.3 & 96.2 $\pm$ 1.3 & 96.1 $\pm$ 1.1 & 96.5 $\pm$ 1.2 \\ \hline
Ours (LR) & 79.5 $\pm$ 1.0 & 28.0 $\pm$ 2.2 & 84.3 $\pm$ 1.7 & 50.0 $\pm$ 4.0 & 91.0 $\pm$ 0.9 & 92.2 $\pm$ 0.6 & 92.6 $\pm$ 0.8 & 92.3 $\pm$ 0.7 \\
Ours (SVM) & 85.9 $\pm$ 0.7 & 43.6 $\pm$ 2.2 & 82.4 $\pm$ 1.1 & 46.9 $\pm$ 3.2 & 91.6 $\pm$ 0.7 & 92.4 $\pm$ 0.6 & 93.0 $\pm$ 0.9 & 92.6 $\pm$ 0.8 \\
Ours (RF) & 88.2 $\pm$ 0.7 & 49.5 $\pm$ 2.0 & 83.6 $\pm$ 0.9 & 49.0 $\pm$ 1.6 & 95.3 $\pm$ 0.7 & 96.8 $\pm$ 0.5 & 95.3 $\pm$ 0.8 & 96.0 $\pm$ 0.7 \\
Ours (XGBoost) & \textbf{90.0 $\pm$ 0.6} & \textbf{54.8 $\pm$ 2.6} & \textbf{85.7 $\pm$ 0.9} & \textbf{52.3 $\pm$ 2.2} & \textbf{97.2 $\pm$ 0.6} & \textbf{97.9 $\pm$ 0.4} & \textbf{97.3 $\pm$ 0.4} & \textbf{97.6 $\pm$ 0.4} \\ \hline
\end{tabular}
}
\caption{\label{tab:p19-p12-pam_results} Performance comparison of our method with various state-of-the-art deep learning methods across three datasets (P19, P12, and PAM).}
\end{table*}

On the MIMIC-III Benchmarks – mortality prediction subset, we similarly compare our method against GRU-D, IP-Net, Phased-LSTM, Transformer, Latent-ODE, and SeFT. The performance of these methods is reported by the authors of SeFT, who use the same benchmarking cohort as ours from the larger MIMIC-III database, making the results more directly comparable. Table \ref{tab:mimic3_results} shows a competitive AUROC score for our method (with the XGBoost classifier) compared to the state-of-the-art GRU-D. A higher AUPRC score indicates that our method performs better on this highly imbalanced dataset.

\begin{table}[ht]
\centering
\renewcommand{\arraystretch}{1.2}
\setlength{\tabcolsep}{8pt}
\begin{tabular}{l|cc}
\hline
Methods & \multicolumn{2}{c}{MIMIC-III}   \\ \cline{2-3}   & AUROC                   & AUPRC         \\ \hline
GRU-D                 & 85.7 $\pm$ 0.2          & 52.0 $\pm$ 0.8 \\
GRU-SIMPLE            & 82.8 $\pm$ 0.0          & 43.6 $\pm$ 0.4         \\
IP-NETS               & 83.2 $\pm$ 0.5          & 48.3 $\pm$ 0.4         \\
PHASED-LSTM           & 80.3 $\pm$ 0.4          & 37.1 $\pm$ 0.5         \\
TRANSFORMER           & 82.1 $\pm$ 0.3          & 42.6 $\pm$ 1.0         \\
LATENT-ODE$^{\dagger}$ & 80.9 $\pm$ 0.2          & 39.5 $\pm$ 0.5         \\
SEFT-ATTN             & 83.9 $\pm$ 0.4          & 46.3 $\pm$ 0.5         \\ \hline
Ours (LR)         & 83.4 $\pm$ 0.6 & 45.8 $\pm$ 1.6 \\
Ours (SVM)         & 78.3 $\pm$ 1.4 & 45.5 $\pm$ 1.3 \\
Ours (RF)         & 83.9 $\pm$ 0.8 & 50.0 $\pm$ 1.1 \\
Ours (XGBoost)         & \textbf{85.9 $\pm$ 0.8} & \textbf{53.6 $\pm$ 1.2} \\ \hline
\end{tabular}
\caption{\label{tab:mimic3_results} Performance comparison of our method with various deep learning methods on MIMIC-III Benchmarks: mortality prediction subset.}
\end{table}

\subsection{Baseline Methods with Raw Input}
To validate that the feature extraction process is the key contributor to the performance of our proposed method, rather than the capabilities of the prediction models themselves, we design a set of comparative experiments. These experiments utilize the same types of prediction models (e.g., XGBoost and LR), but take the raw, high-dimensional time-series data as input, without applying dimensionality reduction. By comparing the performance, we assess the contribution of feature extraction.

Since the raw data contains unobserved missing values (NaNs), and the prediction models we use (except for XGBoost, which handles missing values internally) require complete input data, imputation is needed. We employ various imputation strategies to fill in the missing values in the time series variables. They are:

\begin{itemize}
    \item \textbf{Mean Imputer}: all NaN time series variable values are replaced with the mean of the observed values for the same variable throughout the training split of the dataset.
    
    \item \textbf{Forward Imputer}: this method forwards the last observed value within each time series to fill NaN values. If no previous value is available, the mean of the same variable across the training split is used instead, ensuring that missing initial values are assigned a default.
    
    \item \textbf{Linear Imputer}: missing values are filled by drawing a linear connection between the nearest observed time-value pairs before and after the NaN. If only one observed value is available for a certain variable in the time series, all NaN values for that variable are replaced with that observed value. When no value is observed, the mean always serves as a fallback.
\end{itemize}

We stack the imputed variable array $\boldsymbol{X}$ with the time stamp vector $\boldsymbol{t}$ as a full feature array $\boldsymbol{F}_\text{full} \in \mathbb{R}^{L \times (D+1)}$ to represent each raw time series in the dataset. Since the prediction models we use all require fixed-size input vectors, we pad each feature array into the same shape $L_\text{max} \times (D+1)$, where $L_\text{max}$ is the maximum sequence length in the training split, by prepending $0$'s to each of the variable columns and the time stamp column. During testing, if a test sample has a sequence length larger than $L_\text{max}$, we trim the beginning time steps to make it shorter. Eventually, $\boldsymbol{F}_\text{full}$ is normalized and flattened before being taken as input by XGBoost or LR.

\subsection{Comparison to Baseline Methods with Raw Input}

We extensively compare the predictive power of our summary features with a set of raw input and imputation-based baselines. Tables \ref{tab:p19-p12-mimic3_comp} and \ref{tab:pam_comp} show the results on all four datasets examined.

Across the P12, MIMIC-III, and PAM datasets, the classification head (whether a logistic regressor or XGBoost) achieves the best or comparable performance when using the statistical features extracted by our proposed method. Notably, XGBoost alone, when applied to raw input, fails to surpass previous state-of-the-art deep learning models. This result suggests that our feature extraction approach not only effectively preserves the discriminative information essential for classification but also enhances its expressiveness, thus benefiting the downstream classification head.

Please note that P19 appears to be an outlier in relation to the previous findings. We argue that this is due to the inherent properties of the P19 dataset itself, which will be further examined in the next section.

\begin{table*}[ht]
\centering
{\fontsize{9}{12}\selectfont
\renewcommand{\arraystretch}{1}
\setlength{\tabcolsep}{1pt}
\begin{tabular}{l|cc|cc|cc}
\hline
Methods & \multicolumn{2}{c|}{P19} & \multicolumn{2}{c|}{P12} & \multicolumn{2}{c}{MIMIC-III} \\ \cline{2-7} 
 & AUROC & AUPRC & AUROC & AUPRC & AUROC & AUPRC \\ \hline
Mean-Imp + LR & 74.4 $\pm$ 1.5 & 43.4 $\pm$ 2.7 & 64.4 $\pm$ 1.3 & 26.9 $\pm$ 2.0 & 70.4 $\pm$ 1.4 & 28.5 $\pm$ 1.5 \\
For-Imp + LR & 83.0 $\pm$ 0.7 & 54.7 $\pm$ 2.3 & 70.7 $\pm$ 2.0 & 33.0 $\pm$ 3.1 & 76.1 $\pm$ 0.7 & 35.1 $\pm$ 0.9 \\
Lin-Imp + LR & 83.4 $\pm$ 0.9 & 55.7 $\pm$ 1.6 & 73.2 $\pm$ 1.8 & 34.2 $\pm$ 2.8 & 76.4 $\pm$ 0.5 & 36.2 $\pm$ 0.9 \\
Proposed + LR & 79.5 $\pm$ 1.0 & 28.0 $\pm$ 2.2 & 84.3 $\pm$ 1.7 & 50.0 $\pm$ 4.0 & 83.4 $\pm$ 0.6 & 45.8 $\pm$ 1.6 \\ \hline
Raw + XGB & \textbf{95.6 $\pm$ 0.3} & \textbf{80.1 $\pm$ 0.7} & 81.1 $\pm$ 1.2 & 44.0 $\pm$ 2.2 & 83.3 $\pm$ 0.9 & 48.2 $\pm$ 2.2 \\
Mean-Imp + XGB & 95.2 $\pm$ 0.3 & 78.5 $\pm$ 0.6 & 81.4 $\pm$ 1.5 & 44.1 $\pm$ 1.8 & 83.3 $\pm$ 0.6 & 48.1 $\pm$ 1.9 \\
For-Imp + XGB & 94.8 $\pm$ 0.4 & 77.8 $\pm$ 1.2 & 85.0 $\pm$ 1.1 & 51.2 $\pm$ 2.0 & 85.7 $\pm$ 0.5 & 53.3 $\pm$ 1.8 \\
Lin-Imp + XGB & 94.0 $\pm$ 0.3 & 75.5 $\pm$ 1.8 & 85.3 $\pm$ 1.2 & 52.2 $\pm$ 2.1 & \textbf{85.9 $\pm$ 0.7} & 53.5 $\pm$ 2.1 \\
Proposed + XGB & 90.0 $\pm$ 0.6 & 54.8 $\pm$ 2.6 & \textbf{85.7 $\pm$ 0.9} & \textbf{52.3 $\pm$ 2.3} & 85.9 $\pm$ 0.8 & \textbf{53.6 $\pm$ 1.2} \\ \hline
\end{tabular}
}
\caption{\label{tab:p19-p12-mimic3_comp} Performance comparison: our proposed statistical features vs. raw input and imputed features; on P19, P12, and MIMIC-III datasets.}
\end{table*}

\begin{table*}[ht]
\centering
\renewcommand{\arraystretch}{1.2}
\setlength{\tabcolsep}{5pt}
\begin{tabular}{l|cccc}
\hline
Methods & \multicolumn{4}{c}{PAM}   \\ \cline{2-5} & Accuracy & Precision & Recall & F1 \\ \hline
Mean-Imp + LR & 54.7 $\pm$ 2.1 & 58.4 $\pm$ 2.7 & 53.3 $\pm$ 2.2 & 52.0 $\pm$ 2.0 \\
For-Imp + LR & 57.2 $\pm$ 1.7 & 62.0 $\pm$ 1.4 & 54.7 $\pm$ 1.2 & 53.6 $\pm$ 1.3 \\
Lin-Imp + LR & 60.1 $\pm$ 2.2 & 63.0 $\pm$ 1.9 & 58.4 $\pm$ 1.6 & 57.8 $\pm$ 1.7 \\
Proposed + LR & 91.0 $\pm$ 0.9 & 92.2 $\pm$ 0.6 & 92.6 $\pm$ 0.8 & 92.3 $\pm$ 0.7 \\ \hline
Raw + XGB & 85.1 $\pm$ 0.7 & 88.9 $\pm$ 0.5 & 84.1 $\pm$ 1.2 & 85.7 $\pm$ 0.9 \\
Mean-Imp + XGB & 85.7 $\pm$ 1.0 & 89.1 $\pm$ 0.9 & 84.6 $\pm$ 1.2 & 86.0 $\pm$ 1.2 \\
For-Imp + XGB & 94.7 $\pm$ 0.8 & 96.0 $\pm$ 0.6 & 94.8 $\pm$ 0.7 & 95.3 $\pm$ 0.6 \\
Lin-Imp + XGB & 94.6 $\pm$ 0.8 & 95.8 $\pm$ 0.7 & 94.9 $\pm$ 0.7 & 95.3 $\pm$ 0.7 \\
Proposed + XGB & \textbf{97.2 $\pm$ 0.6} & \textbf{97.9 $\pm$ 0.4} & \textbf{97.3 $\pm$ 0.4} & \textbf{97.6 $\pm$ 0.4} \\ \hline
\end{tabular}
\caption{\label{tab:pam_comp} Performance comparison: our proposed statistical features vs. raw input and imputed features; on PAMAP2.}
\end{table*}


 

\section{Discussion}


\subsection{A Closer Look at P19's Anomaly}
In the previous section, we observed that our extracted statistical features enhance the discriminative power of both classification heads in the P12, MIMIC-III, and PAM datasets, but not in the case of P19. In this case, our proposed statistical features, which eliminate the temporal axis, are outperformed by both the raw input and the imputed input.

As Table \ref{tab:p19-p12-mimic3_comp} shows, an XGBoost classifier applied to the raw input achieves optimal performance for P19. However, performance degrades with imputation as the imputer introduces more temporal smoothing. This suggests that the missing patterns in the original P19 data contain significant predictive information that helps the XGBoost classifier differentiate between positive and negative instances.

To investigate further, we computed the masking array M (defined in the Methods section) for each time series instance in the P19 dataset and used it to train a classifier. We preprocessed all mask arrays to uniform sequential length by prepending $0$s to each variable channel. Using the same 5-fold cross-validation procedure, we calculated the classification metrics presented in Table \ref{tab:mask_input_results}. We conducted identical experiments in all four datasets, reporting AUROC for P19, P12, and MIMIC-III, and the F1 score for PAM.

For P12, MIMIC-III, and PAM, classifiers using only mask arrays (M + LR and M + XGB) performed significantly worse than those using the original raw input (X) or our proposed statistical features (F). However, P19 presents a striking exception. The mask array, along with XGBoost, achieved an AUROC of 94.2\%, only 1.4\% below the raw input. Moreover, with a logistic regressor, the mask array achieved 85.7\% AUROC, surpassing all three imputation-based methods that used original time series variables before feature extraction. This provides compelling evidence that in the P19 dataset, missing data patterns strongly correlate with sepsis labels, serving as powerful predictors comparable to actual time-series measurements (vital signals and lab results). Nevertheless, our proposed statistical features still outperform state-of-the-art deep learning methods reported in the recent literature.

\begin{table*}[ht]
\centering
\footnotesize
\renewcommand{\arraystretch}{1.2}
\setlength{\tabcolsep}{6pt}
\begin{tabular}{l|c|c|c|c}
\hline
Experiments & P19 AUROC & P12 AUROC & MIMIC-III AUROC & PAM F1 \\ \hline
$\boldsymbol{M}$ + LR & 85.7 $\pm$ 0.6 & 56.4 $\pm$ 2.1 & 60.3 $\pm$ 1.0 & 14.6 $\pm$ 1.0 \\ 
$\boldsymbol{F}$ + LR & 79.5 $\pm$ 1.0 & 84.3 $\pm$ 1.7 & 83.4 $\pm$ 0.6 & 92.3 $\pm$ 0.7 \\ \hline
$\boldsymbol{M}$ + XGB & 94.2 $\pm$ 0.3 & 70.9 $\pm$ 1.0 & 70.2 $\pm$ 1.3 & 8.5 $\pm$ 0.4 \\ 
$\boldsymbol{X}$ + XGB & 95.6 $\pm$ 0.3 & 81.1 $\pm$ 1.2 & 83.3 $\pm$ 0.9 & 85.7 $\pm$ 0.9 \\ 
$\boldsymbol{F}$ + XGB & 90.0 $\pm$ 0.6 & 85.7 $\pm$ 0.9 & 85.9 $\pm$ 0.8 & 97.6 $\pm$ 0.4 \\ \hline
\end{tabular}
\caption{\label{tab:mask_input_results} Extensive experiments on discriminative power of mask array $\boldsymbol{M}$ across four datasets.}
\end{table*}

\subsection{Efficiency of Statistical Features}

This study offers valuable insight into modeling real-world time series data, which are often characterized by irregular sampling intervals and missing values.

Most RNN- or Transformer-based models are built on the core principle of learning meaningful latent representations from observed variables at each time step. However, in real-world scenarios, such as ICU stays that involve vital sign readings and lab test results, the parameter space is typically only partially observed and relatively limited compared to traditional deep learning domains such as image classification. Modeling a patient’s health state at each time point based solely on observed data is neither efficient nor completely accurate, as unobserved factors often play a significant role in influencing the general health state. This inherent limitation increases the likelihood of deep learning models overfitting to noisy fluctuations in the training data, ultimately resulting in biased and less reliable predictions. \revise{In contrast, our statistical approach mitigates this by aggregating temporal information into robust summary features, thereby avoiding overfitting to local irregularities while preserving essential predictive signals.}

Furthermore, training such models through stochastic gradient descent \cite {ruder2016overview} is time-consuming, requiring several minutes per epoch 
\cite{horn2020set}. Deep learning models trained on these datasets require gigabytes of GPU memory to run and take approximately one minute to perform inference on a test dataset \cite{li2024time}. In contrast, our method directly derives all the features from the original data by computing only essential statistical measures. This streamlined approach ensures great efficiency for both training and inference. Preprocessing the data and extracting statistical features require only a single linear pass through the entire dataset. The XGBoost classifier itself contains only tens of thousands of parameters. During inference, each time series instance requires fewer than 1,000 FLOPs to generate a prediction, compared to the hundreds of GFLOPs required by a vision transformer for time series modeling to perform inference on a single instance\cite{li2024time}. The efficiency and interpretability of the entire end-to-end process make it an ideal solution to many practical cases of modeling irregular multivariate time series with missing observations.

\subsection{Informative Missing Patterns}

Previous studies, such as GRU-D, have highlighted that missing patterns in real-world time series data can carry meaningful information and, therefore, require careful handling within machine learning pipelines. Our study demonstrates that by focusing solely on the differences between consecutive observations, regardless of the time intervals between observations and the overall temporal patterns, the classification model can still achieve strong performance by leveraging the statistical features of these differences. Specifically, by calculating the mean and standard deviation of these differences, our statistical features effectively capture both the trend and the rate of change for each predictor variable. \revise{To empirically verify this, we conducted an XGBoost feature importance analysis (using Total Gain) on the P19 dataset, as shown in Fig.~\ref{fig:feature_importance}. Results indicate that features capturing the rate of change---$\mu^{(1)}$ (Mean Change) and $\sigma^{(1)}$ (Change Variability)---are highly discriminative, collectively accounting for 46.2\% of the model's total gain. Notably, $\mu^{(1)}$ alone contributes 26.1\%, surpassing the static observed standard deviation $\sigma^{(0)}$ (23.3\%). This confirms that temporal dynamics are essential for accurate classification.}

\begin{figure}[t]
    \centering
    \includegraphics[width=0.6\textwidth]{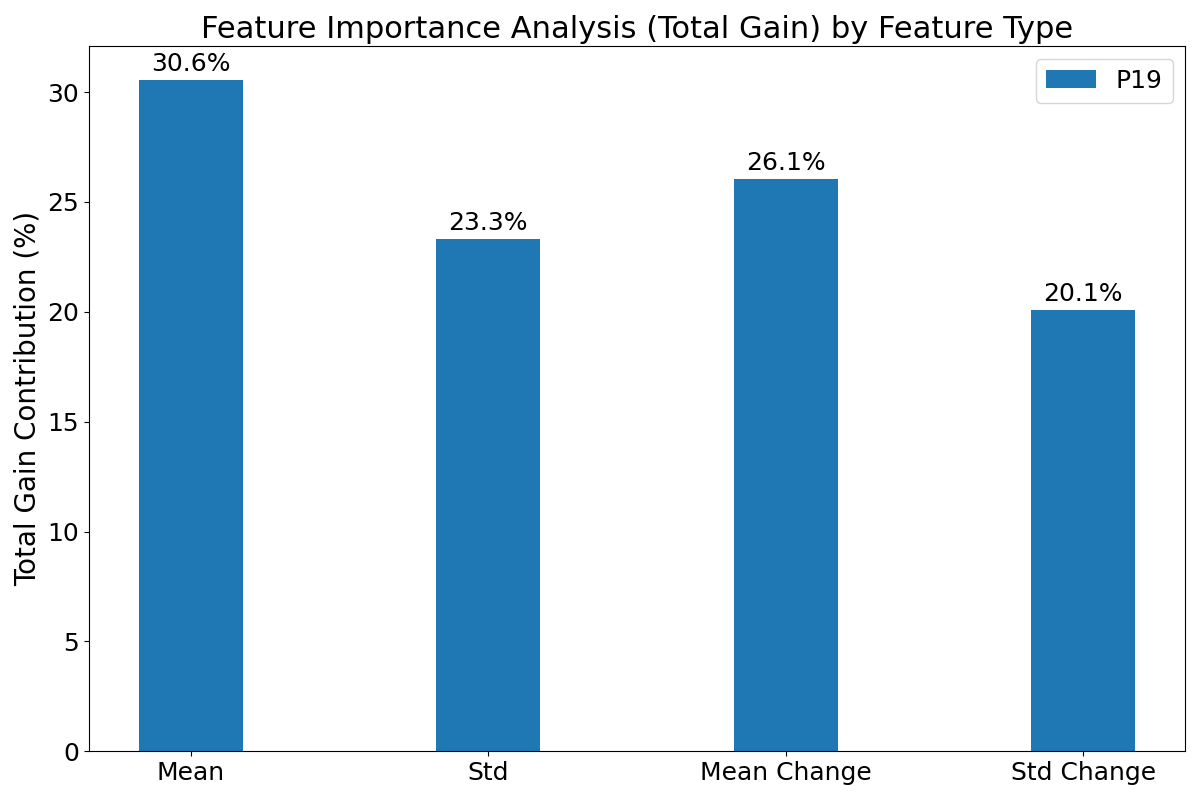}
    \caption{Feature importance analysis on the P19 dataset using XGBoost. The importance metric is Total Gain. For each of the four feature types—$\mu^{(0)}$ (Mean), $\sigma^{(0)}$ (STD), $\mu^{(1)}$ (Mean Change), and $\sigma^{(1)}$ (Change Variability)—we summed the gains of all corresponding features, averaged over 5-fold cross-validation.}
\label{fig:feature_importance}
    \label{fig:feature_importance}
\end{figure}

We argue that, in real-world scenarios, the missing patterns of these variables often correlate with their underlying trends of change. For example, a doctor can monitor a patient more frequently if the patient’s vital signs indicate a dangerous rate or direction of change, leading to fewer missing values for certain lab test results and a reduced pattern of missing data. Therefore, it is not always necessary to fully retain the missing data patterns or explicitly extract features from them. Whether or not to account for missing patterns in detail depends on the specific domain and scope of the data. \revise{As our results in Table 5 demonstrate, this phenomenon is not universal but is a striking exception specific to the P19 dataset. We found compelling evidence that missingness patterns alone are powerful predictors of sepsis; however, this is not the case in the P12, MIMIC-III, and PAM datasets.} Researchers should make model design decisions based on empirical experimental results, rather than incorporating all potential features without careful consideration.

\subsection{Limitation}

\revise{We must acknowledge that our time-agnostic approach is primarily advantageous for endpoint prediction tasks (e.g., predicting overall mortality or sepsis occurrence). A clear limitation of our method is its inability to address tasks requiring high temporal resolution or step-by-step forecasting. Specifically, our method cannot directly address the original objective of the PhysioNet Challenge 2019, which requires identifying the exact hour of sepsis onset. By aggregating temporal information into summary statistics, we trade fine-grained temporal localization for computational efficiency and robustness in endpoint classification. Consequently, our method is best suited for scenarios where the patient's "global state" is the primary interest, rather than the precise timing of future events.}

\section{Conclusion}

Modeling irregular multivariate time series with missing values presents significant challenges for predictive tasks in healthcare and other domains. Although existing approaches often employ complex deep learning architectures focused on temporal interpolation or sophisticated architectural designs, they introduce computational complexity. They may not fully leverage the predictive signals in the data. To address these limitations, we propose a more straightforward yet practical approach that extracts time-agnostic summary statistics—namely, the mean and standard deviation of the observed values, as well as the mean and variability of changes between consecutive observations—to create fixed-dimensional representations that eliminate the temporal axis. Our method, combined with standard classifiers such as logistic regression and XGBoost, achieves state-of-the-art performance across four biomedical datasets (PhysioNet 2012, 2019, PAMAP2, and MIMIC-III), outperforming recent transformer and graph-based models while significantly reducing computational complexity. Through extensive ablation studies, we demonstrate that our performance gains are primarily derived from the feature extraction process rather than classifier choice, and our summary statistics outperform raw and imputed inputs in most benchmarks. In particular, our investigation reveals that missing patterns themselves can encode strong predictive signals, particularly in sepsis prediction, where missing indicators alone achieve very high AUROC. Although our work focuses on biomedical time series, we believe that our approach offers a broadly applicable, efficient, and interpretable solution whenever task objectives permit time-agnostic representations, challenging the necessity of complex temporal modeling for irregular time series classification.

\bibliographystyle{plainnat}
\bibliography{references}

\end{document}